\documentclass{article}
\usepackage{PRIMEarxiv}

\usepackage[utf8]{inputenc} 
\usepackage[T1]{fontenc}    
\usepackage{hyperref}       
\usepackage{url}            
\usepackage{booktabs}       
\usepackage{amsfonts}       
\usepackage{nicefrac}       
\usepackage{microtype}      
\usepackage{lipsum}
\usepackage{fancyhdr}       
\usepackage{graphicx}       
\graphicspath{{media/}}     
\usepackage{multirow}
\usepackage{tikz}
\usetikzlibrary{shapes, arrows.meta, positioning}
\usepackage{algorithm}
\usepackage{algorithmic}
\usepackage{amsmath} 
\usepackage{amssymb} 

\title{Fine-Tuning Cycle-GAN for Domain Adaptation of MRI Images}

\author{
  Mohd Usama$^{\mathrm{\ast \ddag}}$ \\
  \texttt{mohd.usama@umu.se} 
   \And
  Belal Ahmad$^{\mathrm{\dag}}$ \\
  \texttt{belalamu.63@yahoo.com} 
     \And
  Faleh Menawer R Althiyabi$^{\mathrm{\ddag}}$ \\
  \texttt{faleh.thiyabi@kfupm.edu.sa} 
     \And
  \\
  \\
  \textit{$^{\mathrm{\ast}}$Department of Diagnostics and Intervention, and Biomedical Engineering, Umea University, Sweden}\\
    \textit{$^{\mathrm{\dag}}$Department of Computer Science and Information Engineering, National Taipei University of Technology, Taiwan}\\
    \textit{$^{\mathrm{\ddag}}$IRC for Bio Systems and Machines, King Fahd University of Petroleum and Minerals, Saudi Arabia}\\\\
  \scshape A preprint - \today
}

\begin{document}
\maketitle

\begin{abstract}
Magnetic Resonance Imaging (MRI) scans acquired from different scanners or institutions often suffer from domain shifts owing to variations in hardware, protocols, and acquisition parameters. This discrepancy degrades the performance of deep learning models trained on source domain data when applied to target domain images. In this study, we propose a Cycle-GAN-based model for unsupervised medical-image domain adaptation. Leveraging CycleGANs, our model learns bidirectional mappings between the source and target domains without paired training data, preserving the anatomical content of the images. By leveraging Cycle-GAN capabilities with content and disparity loss for adaptation tasks, we ensured image-domain adaptation while maintaining image integrity. Several experiments on MRI datasets demonstrated the efficacy of our model in bidirectional domain adaptation without labelled data. Furthermore, research offers promising avenues for improving the diagnostic accuracy of healthcare. The statistical results confirm that our approach improves model performance and reduces domain-related variability, thus contributing to more precise and consistent medical image analysis.
\end{abstract}


\section{Introduction}
Deep learning technology can provide important information on a diversity of prediction tasks, such as medical imaging, which includes segmentation \cite{litjens2017survey}, precise diagnosis \cite{zhang2019noise} and clinical outcome prediction \cite{yan2019discriminating}. Furthermore, it frequently performs inconsistently when applied to data obtained under diverse scenarios, such as imaging equipment, patient populations, and acquisition protocols \cite{takao2011effect}\cite{usama2026domain}. This imaging heterogeneity can mitigate the generalizability of models because they exhibit inappropriate and perplexing features. In the field of medical imaging, poor generalization exhibits important inadequacies for the extensive clinical adoption of deep learning-based predictors\cite{zhang2020collaborative}\cite{choudhary2020advancing}. GANs are a class of deep learning techniques that employ adversarial training to read an image of the target distribution and have been demonstrated to be very useful for synthesizing image data in various circumstances \cite{goodfellow2020generative}. Advancements in the computer vision community have demonstrated how GANs can be utilized to train deep learning-based models that are robust to adversarial trepidations \cite{wang2019direct}\cite{usama2020self}\cite{guan2021domain}. In the context of modelling natural variability, Robey et al. \cite{robey2020model} utilized comprehensible variation models to train vigorous deep learning against issues such as weather conditions in lane sign identification and background color in digit identification. Despite these improvements, difficulties remain in processing multisite data and generalizing predictors to new sites in the field of medical imaging. Moreover, several deep learning-based techniques for image harmonization have been presented\cite{chen2017disease}\cite{chen2017deep}\cite{yan2015deep}. 

Generative Adversarial Networks (GANs)\cite{creswell2018generative} have demonstrated remarkable achievements in image generation, editing, and representation learning\cite{xie2020mi}\cite{zhang2022c}. These methods leverage the concept of adversarial loss, driving the generated images toward indistinguishability from authentic images. This approach has proven particularly effective for image generation, which aligns with the primary objective of computer graphics optimization research. Recent advancements have witnessed the application of GANs to conditional image generation tasks, such as text-to-image translation, image inpainting, and future prediction, extending their capabilities to domains including videos and 3D data representation. The adversarial loss strategy remains a cornerstone of GANs' success of GANs, enabling them to produce highly realistic and visually captivating content. Zili yi et.al \cite{yi2017dualgan} proposed a novel dual-GAN framework that facilitates the training of image translation models using two sets of unlabeled images from distinct domains. His approach is based on the idea of using two GANs, one for each domain, to learn the mapping between the two domains. The first GAN is trained to transform images from the first to the second domain, whereas the second GAN is trained to convert images from the second to the first domain. By jointly training the two GANs, they can learn a mapping that is both accurate and invertible. Furthermore, the results indicate that the DualGAN-based approach can expressively enhance the outputs of GAN for several image-to-image interpretation tasks. However, his method was outclassed by the conditional GAN (C-GAN) for particular tasks that involved semantic-based rules. A novel method proposed by Zhang \cite{zhang2019noise} utilized the noise-adaptation GAN approach, which comprised a generator and two discriminators. The purpose of the generator is to learn the data source from the source domain to the objective domain. Of the two discriminators, one ensures that the generated images possess the same noise patterns as those in the target domain, whereas the second guarantees that the content within the generated images remains intact. The results showed that the proposed approach was able to attain the contents while transmitting the noise styles from the source to the target domain. Furthermore, noise style transfer improves the accuracy of subsequent analytical tasks such as segmentation and classification. A selective feature network was proposed by Tan et al. \cite{tan2022selective} to reduce LDCT image noise using Cycle-GAN. The issue of potential mode collapse in GAN and its impact on clinical diagnosis accuracy has been examined. In addition, investigating the scalability of the SKF-Cycle-GAN across different medical imaging modalities and datasets could shed light on its versatility and applicability in various clinical scenarios. Furthermore, conducting comparative studies with traditional image enhancement techniques commonly used in medical imaging could provide a comprehensive evaluation of the performance of the SKF-Cycle-GAN and its potential superiority in enhancing image quality for diagnostic use. Moreover, analyzing the computational efficiency of the SKF-Cycle-GAN in processing large-scale medical imaging datasets could help determine its practicality for real-time clinical applications and its integration into existing healthcare systems. Zhang et al. (2019) \cite{zhang2019skrgan} proposed a sketching-rendering GAN (SkrGAN), an unconditional generative adversarial network designed to synthesize realistic medical images from random noise without requiring paired training data. The framework introduces a two-stage generation process: Sketching stage – captures the structural and anatomical outlines of medical images using a coarse generator. Rendering stage – This stage refines the sketches by adding texture, intensity, and fine-grained details to produce high-quality realistic medical images. The authors emphasized that medical image synthesis differs from natural image generation because of its high structural consistency and low texture diversity. SkrGAN addresses this by decomposing the generation process into hierarchical structural and textural models, thereby improving anatomical plausibility. The authors introduced \cite{mahboubisarighieh2024assessing} a 3D dual ‐ cycle-GAN architecture to synthesize multiple MRI contrasts (T1c, T2, and FLAIR) from volumes weighted with T1 data. Trained on BraTS 2021 data, this method leverages both supervised alignment and unpaired samples via two interconnected cycle-GANs sharing weights, balancing the adversarial and cycle-consistency objectives. The only limitation of this study was the generalization of paired versus unpaired data: impact generalization—only ~20\% paired data, so the robustness of the unseen domain is limited. The 3D Dual-Cycle GAN model demonstrated potential for clinical deployment in scenarios with incomplete MRI sequences.

However, the above studies focused on mitigating slight variations in medical images owing to the variability of parameter settings or image collection from different machines that lie in the background pixels of images. In addition, the effectiveness of some of the above studies is bounded by aspects such as the requirement that the training data contain paired subjects who have been imaged at both domain sites \cite{nath2019inter}\cite{zhang2022c2}\cite{zhao2022semantic}, which is very complex to meet in practice with appropriately large and constantly updated training datasets. We propose a modified version of CycleGAN \cite{zhu2017unpaired}, which is a subfield of GANs. Cycle GAN is an approach that automatically trains image-to-image transformation models without requiring paired data. This is accomplished by learning a plot from data in the source domain to the reference target domain, whereas the transformed data maintain and preserve their original semantic information by enforcing identity mapping to the original domain. However, the use of a conventional Cycle-GAN for unsupervised domain adaptation in typical MRI data may not be sufficient to reserve essential structures in the images, which is significant for medical segmentation and other clinical tasks.

The remainder of this paper is organized as follows. Section II presents the materials and methods used in this study. Section III presents the experimental results of this study. Section IV elaborates on the results and discussion of the study. Finally, Section V concludes the paper and outlines future work.

\section{Material and Methods}

\begin{figure*}
\centering
{\includegraphics[scale=0.60]{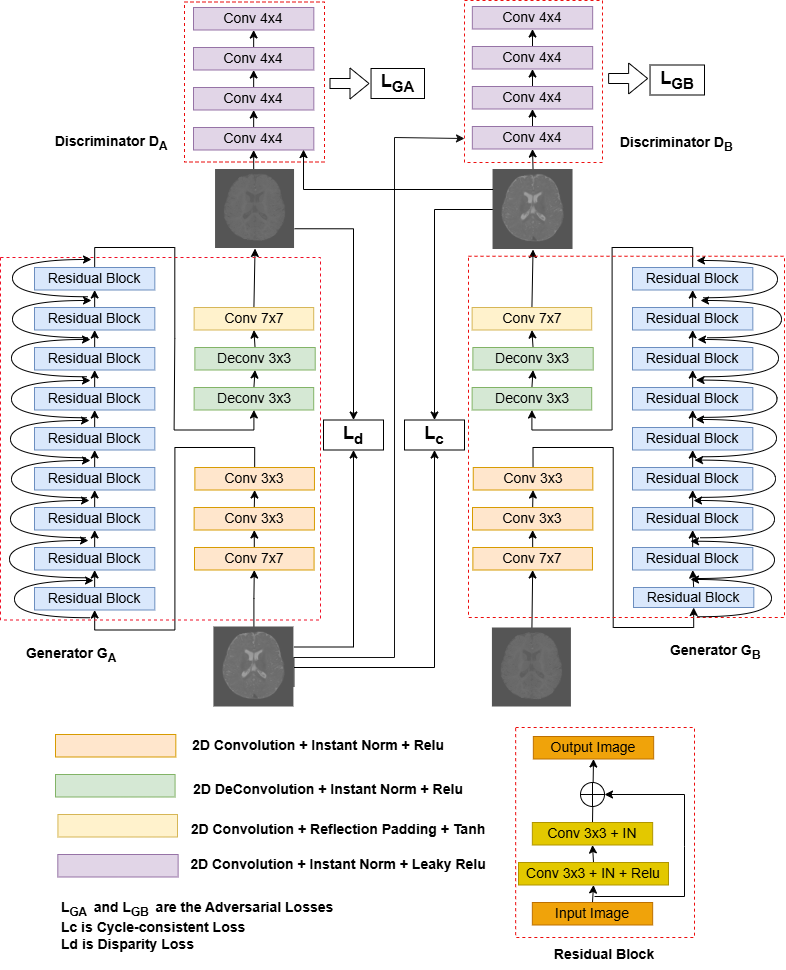}}
\caption{a) Architecture of the Proposed GAN. $G_A$ and $G_B$ are denotes the generator network and $D_A$ and $D_A$ are two discriminators, trained by adversarial losses $\mathcal{L}_{GA}$ and $\mathcal{L}_{GB}$, respectively. $x \sim T1$ denotes the image set in the source domain. $y \sim T2$ represents an image set from the target domain. $L_c$ and $L_d$ indicates the cycle consistent and disparity loss, respectively.}
\label{Figure1}
\end{figure*}

Many studies have been conducted on the domain adaptation of medical images \cite{usama2025domain}, focusing on removing or translating noise in medical images from one domain to another domain. The proposed work focuses on adapting the T1 modality of MRI images into a T2 modality and vice versa. Both T1 and T2 are clean images but differ in the content shades collected from the system with different settings. The details of both image modalities are described in the Dataset subsection. Fig 1. Illustrates the working mechanism of the fine-tuned Cycle-GAN model.\\

The original Cycle-GAN model utilizes two GAN architectures with adversarial and cycle-consistency losses. We modified it by adopting two GAN architectures and optimizing the objective function, including the adversarial and cycle-consistency losses, together with the disparity loss, which is calculated from the difference between the real image of one domain and the corresponding generated image of another domain, and vice versa. This is because both domain images are different in terms of content shade rather than any typical noise. Here, both the cycle-consistent loss and disparity loss contribute to translating the shades of the color region from one domain to another and preserving the structural content of images from one domain to another.\\

$G_A$ and $G_B$ are the generator architectures of the proposed GAN model used to translate the image modality from T1 to T2 and from T2 to T1, respectively. DA and DB are the discriminator architectures of the proposed GAN model used to distinguish the modality between T1 and generated T2 and T2 and generated T1, respectively. Let X  and Y be the image sets from modalities T1 and T2, respectively. During adversarial training, generators GA and GB produce outputs similar to targets Y and X, respectively. The discriminators $D_A$ and $D_B$ distinguish images between X and the translated image G(Y), and Y and the translated image G(X), respectively. Let $\mathcal{L_A}$  and $\mathcal{L_B}$ be the losses for adversarial training of the generators and discriminators, respectively. The typical adversarial losses for both GAN are defined as

\begin{align}
\mathcal{L}_{GA} &= 
\mathbb{E}_{x \sim p_{\text{data}}(x)}\big[\log D_A(x)\big] +
\mathbb{E}_{y \sim p_{\text{data}}(y)}\big[\log\big(1 - D_x(G_A(y))\big)\big], \\[6pt]
\mathcal{L}_{GB} &= 
\mathbb{E}_{y \sim p_{\text{data}}(y)}\big[\log D_B(y)\big] +
\mathbb{E}_{x \sim p_{\text{data}}(x)}\big[\log\big(1 - D_y(G_B(x))\big)\big].
\end{align}

The cycle consistency loss between the domain images is defined as
\begin{equation}
\mathcal{L}_{c}(x, y) = \lvert X_{\text{real}} - X_{\text{gen}} \rvert 
+ \lvert Y_{\text{real}} - Y_{\text{gen}} \rvert
\end{equation}

The corresponding disparity losses between the domain images are calculated as follows:
\begin{equation}
\mathcal{L}_{d}(x, y) = 
\lvert X_{\text{real}} - Y_{\text{gen}} \rvert 
+ \lvert Y_{\text{real}} - X_{\text{gen}} \rvert
\end{equation}

Adversarial losses, together with cycle-consistent and disparity losses, focus on training GANs for the translation of the ROI in the image from T1 to T2 and the preservation of semantic information of images. Finally, our aim is to optimize the objective function by combining all the losses defined as

\begin{equation}
\mathcal{L}_{G,D,X,Y} =
\mathcal{L}_{GA} + \mathcal{L}_{GB}
+ \lambda_{1}\,\mathcal{L}_{c}(x, y)
+ \lambda_{2}\,\mathcal{L}_{d}(x, y)
\end{equation}

where $\lambda_{1}$ and $\lambda_{2}$ are constant hyperparameters.

\begin{figure*}
\centering
{\includegraphics[scale=0.90]{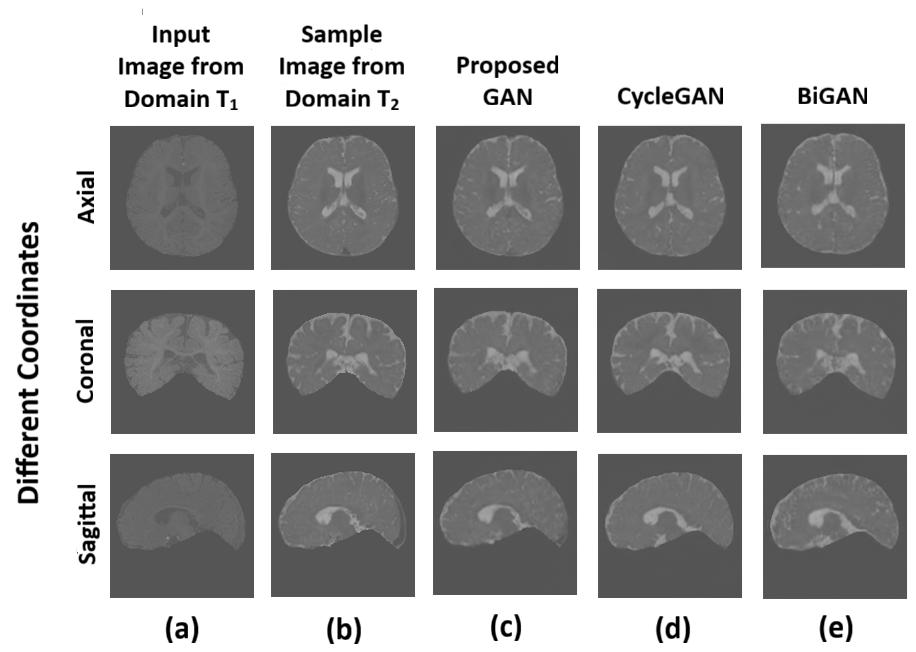}}
\caption{Results show MRI adaptation of T1 modality into T2 by Proposed GAN, CycleGAN, BiGAN, and DualGAN from three different coordinates. Row one two and three shows results from Axial, Coronal, and Sagittal coordinates, respectively.}
\label{Figure2}
\end{figure*}

\begin{figure*}
\centering
{\includegraphics[scale=0.90]{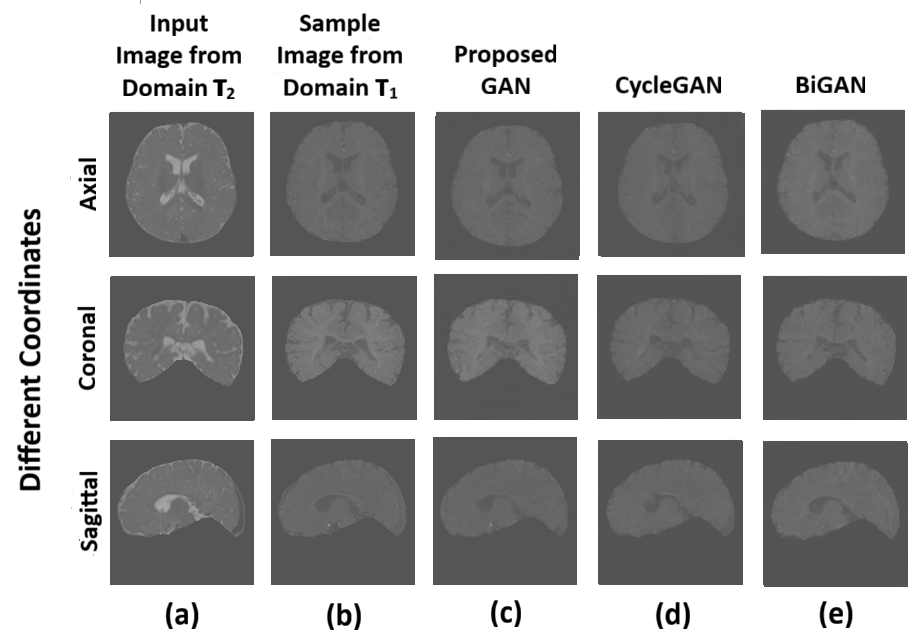}}
\caption{Results show MRI adaptation of T2 modality into T1 by Proposed GAN, CycleGAN, BiGAN, and DualGAN from three different coordinates. Row one two and three shows results from Axial, Coronal, and Sagittal coordinates, respectively.}
\label{Figure2}
\end{figure*}

\subsection{Dataset}
To evaluate the effectiveness of fine-tuning Cycle-GAN for MRI domain adaptation, we conducted experiments on the iSeg 2017 dataset of infant brain MRI. This dataset consists of T1 and T2 modalities \cite{wang2019benchmark}. Both modalities were recorded using varying machine parameter settings.  T1 images were recorded with 144 sagittal slices with parameters of resolution = 1 × 1 × 1 mm3, TR/TE = 1900/4.38 ms, and flip angle = 7◦, whereas T2 images were recorded with 64 axial slices with parameters of resolution = 1.25x1.25x1.95 mm3, TR/TE = 7380/119 ms, and flip angle = 150◦. Fig. 1 depicts both T1 and T2 images of a 6-month infant subject from three coordinates: sagittal, coronal, and axial views. The images were differentiated into white matter, gray matter, and cerebrospinal fluid. The iSEG-2017 organizers prepared this dataset using the iBEAT tool. Images from all three coordinates were used to train and test the model for T1 to T2 adaptation and vice versa.

\subsection{Domain Definition}
\textbf{Source and Target Domain:} The Iseg2019 dataset was employed as the source domain. It contains pre-processed multimodal brain MRIs with four modalities (T1, T1c, T2, and FLAIR). For this study, only T1- and T2-weighted images were used to ensure consistency across domains. T1-weighted images were considered as the source domain, and T2-weighted images were considered as the target domain. A total of 3640 2D images were used for training after converting 10 3D images, among which 1820 were from the T1 modality for the source domain and 1820 were from the T2 modality for the target domain. Similarly, 150 (75 from each modality) were used to test the model. All images were cropped to 144 × 144 for 2D slice-based training.

\subsection{Fine tunning strategy}
We proposed a lightweight fine-tuning protocol that adapted a Cycle-GAN model from scratch. We fine-tuned and trained the model to learn the mapping from the source to the target domain using unpaired image data. The strategy includes:

\textbf{Training from Scratch:} The Cycle-GAN was first trained from scratch on the source domain images for 180 epochs using unpaired image slices.

\textbf{Fine-tuning Initialization and Training Parameters:} The weights of both generators and discriminators were retained, but during early fine-tuning (first 10 epochs), the discriminators were frozen to stabilize adaptation. Optimizer: Adam ($\beta_{1} = 0.5$, $\beta_{2} = 0.999$), Learning Rate: $2*10^{-4}$ during training; linearly decayed to 0 over fine-tuning, Batch Size: 1 (to preserve memory and spatial detail).

\textbf{Disparity Loss}  $\mathcal{L_d}$ was incorporated with cycle-consistent loss during training using features extracted from a Resnet-15 generator architecture to preserve structural integrity across domains.

\section{Results}

For the domain adaptation results, we experimented with multiple GAN models, including BiGAN\cite{donahue2016adversarial}, CycleGAN\cite{zhu2017unpaired}, and fine-tuned CycleGAN with the benchmark dataset ISeg 2019\cite{wang2019benchmark}. In addition, the results were evaluated in multiple ways for robust analysis, such as visual quality assessment, structure similarity index measure (SSIM) \cite{brunet2011mathematical} of images between one domain and another, Bhattacharya distance, and histogram correlation \cite{bhattacharyya1943measure} values to measure the image translation quality between one domain and another.  

\subsection{Visual Quality Assessment}
Visual inspection of the translated MRI images indicated that the fine-tuned CycleGAN model produced images that were not only visually similar to the target domain but also retained critical anatomical structures such as ventricles, gray/white matter boundaries, and lesions. Figures 2 and 3 show side-by-side comparisons of the input/source, target, and translated/reconstructed images, respectively. Compared with the baseline Cycle-GAN and BiGAN trained from scratch, our fine-tuned proposed model exhibited improved tissue contrast and fewer hallucinated artifacts. Figure 2 shows the translation of the T1 modality into T2, whereas Figure 3 shows the translation of the T2 modality into T1 from all three coordinates (axial, coronal, and sagittal) using BiGAN, CycleGAN, and the proposed GAN.

\subsection{Structural Preservation and Cross-Domain Consistency}

\begin{figure*}
\centering
{\includegraphics[scale=0.50]{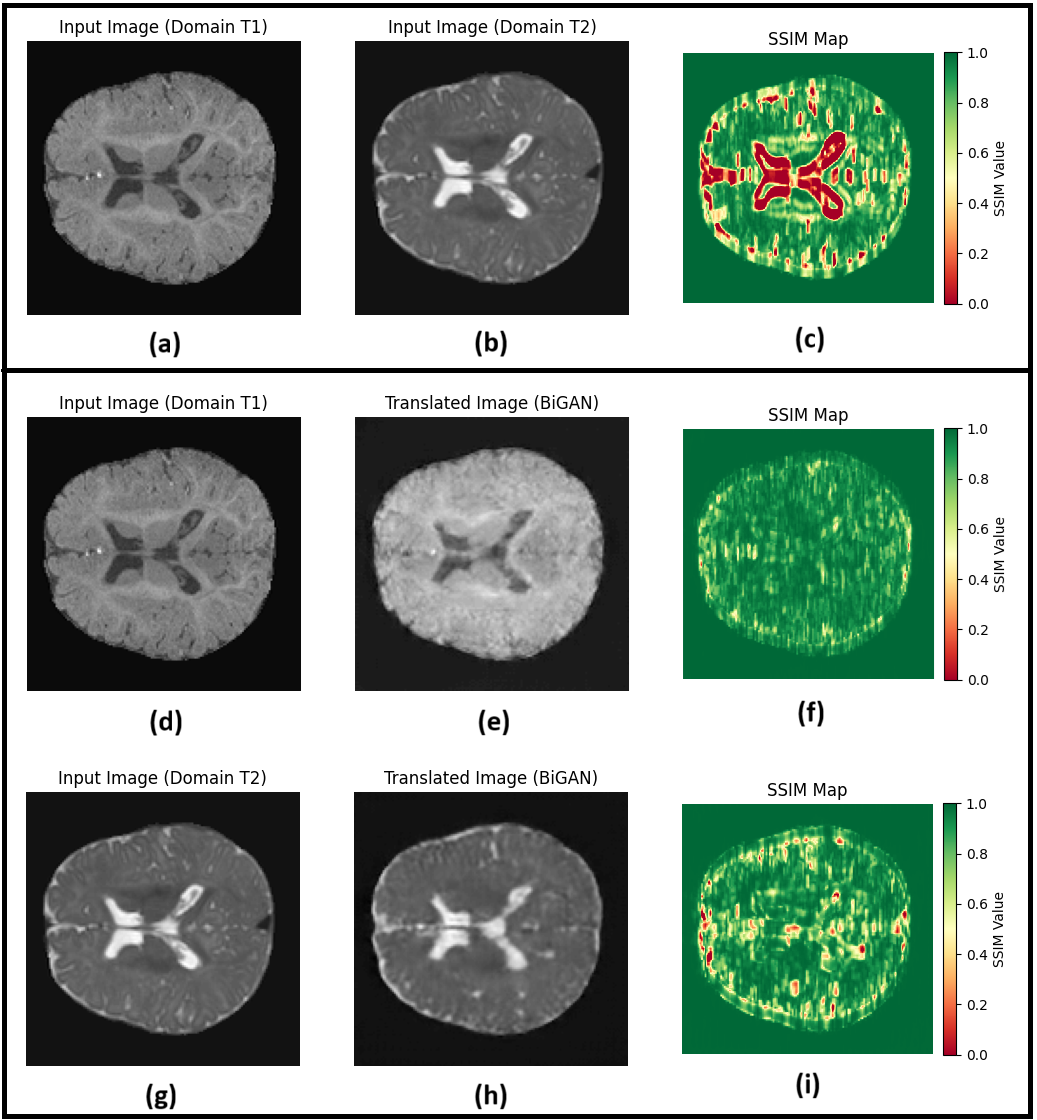}}
\caption{SSIM map from original and generated T1 and T2 image by BiGAN}
\label{Figure3}
\end{figure*}

\begin{figure*}
\centering
{\includegraphics[scale=0.50]{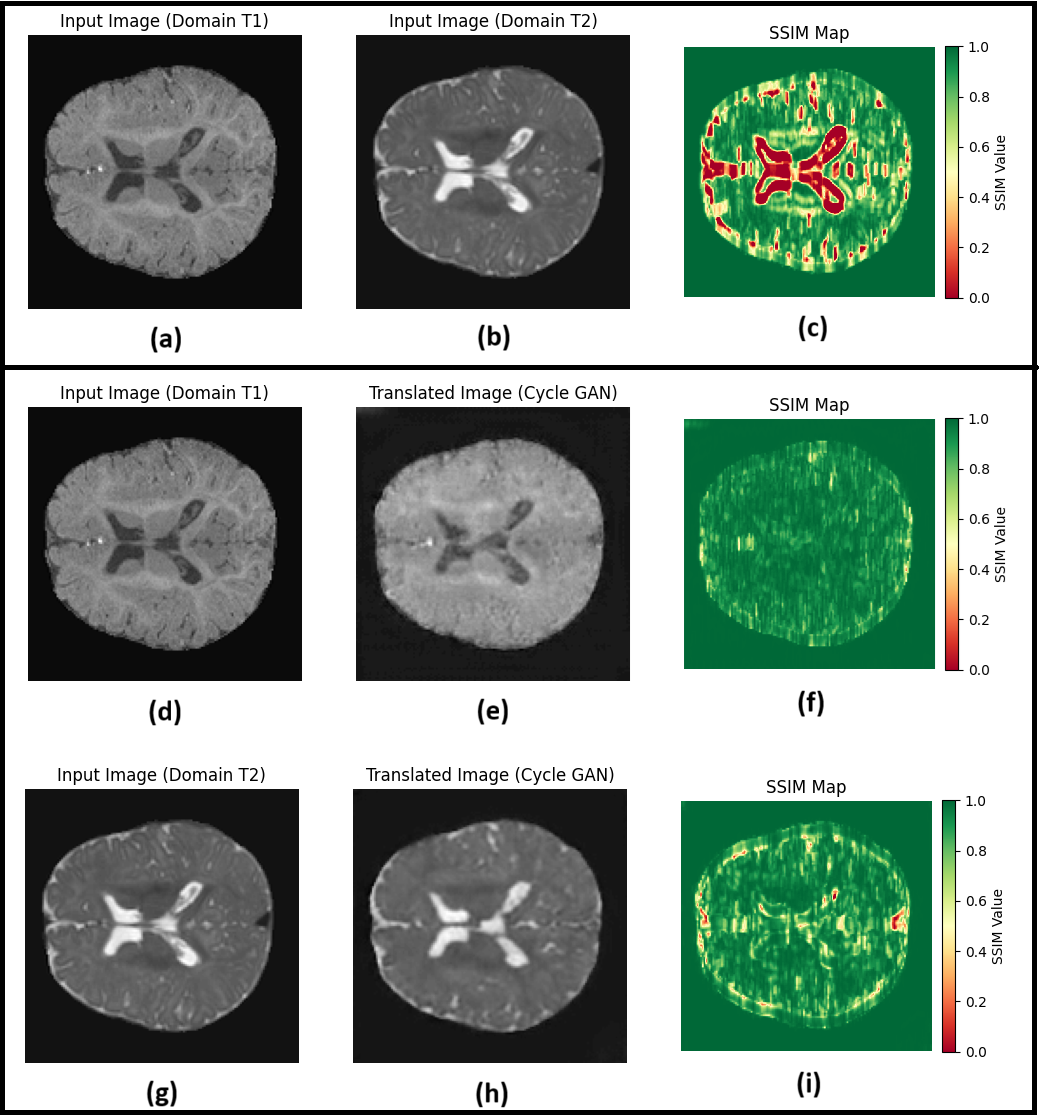}}
\caption{SSIM map from original and generated T1 and T2 image by CycleGAN}
\label{Figure3}
\end{figure*}

\begin{figure*}
\centering
{\includegraphics[scale=0.50]{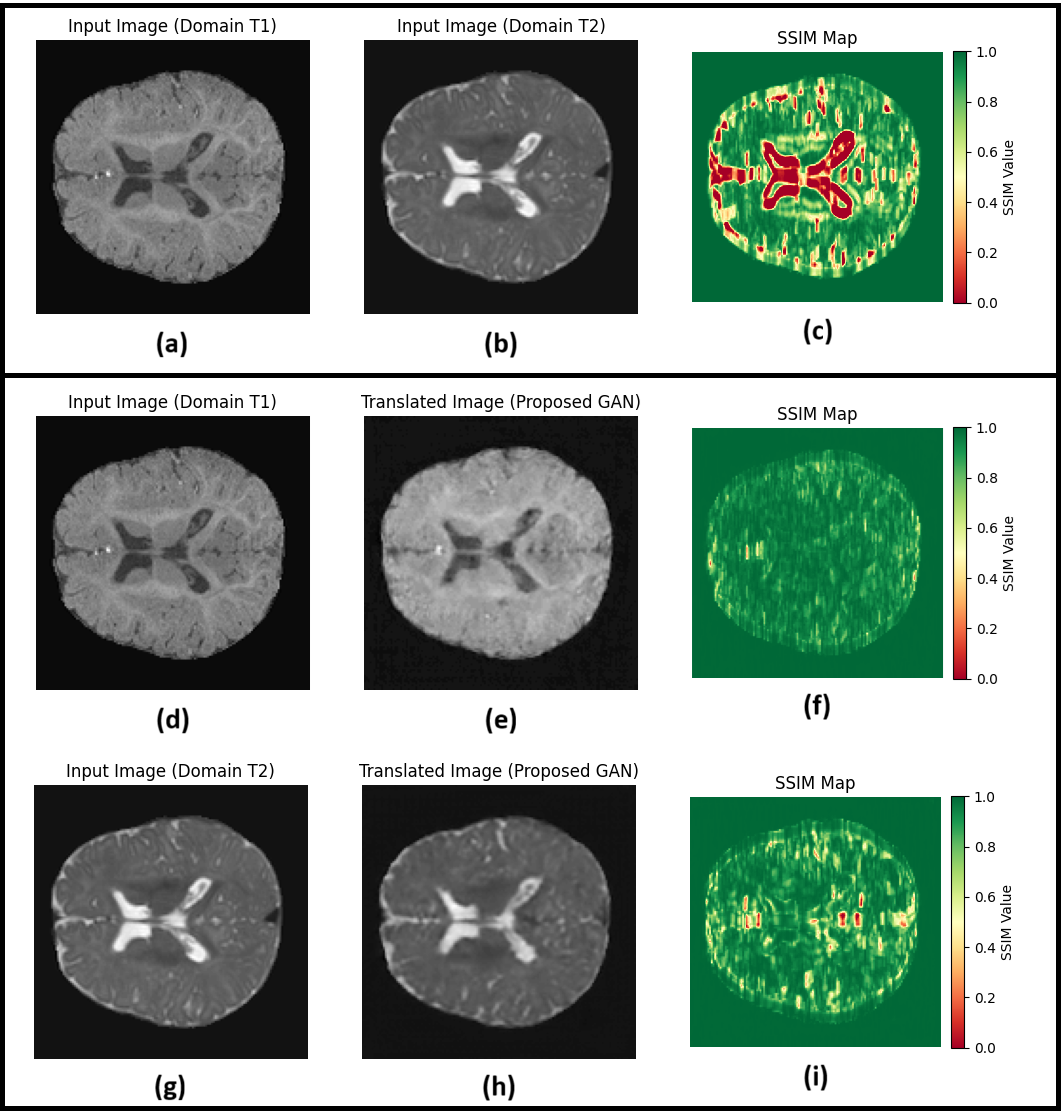}}
\caption{SSIM map from original and generated T1 and T2 image by proposed GAN}
\label{Figure3}
\end{figure*}

Fig 4, 5, and 6 represent the analyses of domain adaptation using the structure similarity index measure (SSIM) between images of different domains. Fig 4(c), 5(c), and 6(c) represent the SSIM between the T1 and T2 modalities before translation. Fig 4(f) and Fig 4(i) represent the SSIM measure between T1 and translated T1, and T2 and translated T2, respectively, using BiGAN. It can be clearly seen that the structured similarity increased in 4(f) and 4(i) compared to 4(c). Furthermore, structural similarity was higher when translating T2 into T1 (SSIM value 0.945, fig 4(f)) than when translating T1 into T2 (SSIM value 0.909, fig 4(i)).  Similarly, Fig 5(f) and 5(i) represent the SSIM measures between T1 and translated T1 and T2 and translated T2, respectively, using CycleGAN. It can be clearly seen that the structured similarity increased in Figs.5(f) and 5(i) compared to that in Fig.5(c). Furthermore, structural similarity was higher when translating T2 into T1 (SSIM value 0.960, fig 5(f)) than when translating T1 into T2 (SSIM value 0.934, fig 5(i)). Similarly, Fig 6(f) and 6(i) represent the SSIM measures between T1 and translated T1 and T2 and translated T2, respectively, using the proposed GAN. It is evident that the structured similarity increased in 6(f) and 6(i) compared to 6(c). Furthermore, the structural similarity was higher when translating T2 into T1 (SSIM value 0.960, fig 6(f)) than when translating T1 into T2 (SSIM value 0.933, fig 6(i)). Despite these differences, both translated modalities demonstrate substantial adaptation of images from one domain to another. The fine-tuning process, augmented by disparity and cycle consistency loss, enabled the model to maintain structural fidelity and anatomical realism, which was evident in visual assessments and structural similarity. 

The models demonstrated effective bidirectional translation between the T1 and T2 domains. Despite the inherent modality differences, such as contrast and intensity variations, the generated images successfully emulated the target domain characteristics while maintaining consistency with the source domain structure ( Figures 2 and 3).

These results highlight the utility of lightweight fine-tuning for domain adaptation in medical imaging. Moreover, the proposed model outperforms BiGAN and the default CycleGAN by preserving core structures better while aligning domain characteristics, which supports a robust cross-modality translation.

\subsection{Quantitative Evaluation and Intra-Modality Consistency}

\begin{table*}[!ht]
\caption{Bhattacharyya distance (BD) and Histogram correlation (HC) values between T1 and Translated T1(from T2)}
\begin{center}
\setlength{\tabcolsep}{5pt}
\begin{tabular}{|c|c|c|}
\hline
\rule{0pt}{1.0\normalbaselineskip}
\rule{0pt}{1.0\normalbaselineskip}
 Method &  Bhattacharyya distance & Histogram correlation \\[5pt]
\hline
\newcommand\Tstrut{\rule{0pt}{2.6ex}}
\rule{0pt}{1.0\normalbaselineskip}
No Translation T1 vs T1 & 0 & 1 \\[5pt]
No Translation T2 vs T2 & 0 & 1 \\[5pt]
No Translation T1 vs T2 & 0.288 & 0.915 \\[5pt]
BiGAN      & 0.274 & 0.965 \\[5pt]
CycleGAN      & 0.143 & 0.970 \\[5pt]
Proposed GAN & 0.011 & 0.997 \\[5pt]
\hline
\end{tabular}
\end{center}
\label{tab5}
\end{table*}

\begin{table*}[!ht]
\caption{Bhattacharyya distance (BD) and Histogram correlation (HC) values between T2 and Translated T2(from T1)}
\begin{center}
\setlength{\tabcolsep}{5pt}
\begin{tabular}{|c|c|c|}
\hline
\rule{0pt}{1.0\normalbaselineskip}
\rule{0pt}{1.0\normalbaselineskip}
 Method &  Bhattacharyya distance & Histogram correlation \\[5pt]
\hline
\newcommand\Tstrut{\rule{0pt}{2.6ex}}
\rule{0pt}{1.0\normalbaselineskip}
No Translation T1 vs T1 & 0 & 1 \\[5pt]
No Translation T2 vs T2 & 0 & 1 \\[5pt]
No Translation T2 vs T1 & 0.288 & 0.915 \\[5pt]
BiGAN      & 0.195 & 0.771 \\[5pt]
CycleGAN      & 0.152 & 0.795 \\[5pt]
Proposed GAN & 0.078 & 0.889 \\[5pt]
\hline
\end{tabular}
\end{center}
\label{tab5}
\end{table*}

To quantify the visual improvements and domain alignment, we performed quantitative assessments using two key statistical metrics: Bhattacharyya Distance (BD) and Histogram Correlation (HC)[31]. These metrics provide insights into the similarity of intensity distributions between the original and translated images, thereby reflecting the quality and realism of the domain-adaptation process. The results are summarized in Tables 1 and 2, respectively.

No Translation cases (T1 vs. T1 and T2 vs. T2) served as baselines and yielded perfect BD and HC scores of zero and one, respectively, confirming that the metric calculations were correctly aligned with the ground truth comparisons. These results validate that the metrics are sensitive to the changes introduced by the translation. We evaluated the BD and HC between T1 and translated T2 and T2 and translated T1 for the three models (BiGAN, CycleGAN, and the proposed GAN). From Table 1, it is clear that BiGAN and CycleGAN achieve low BD (0.274 and 0.143, respectively) and high HC (0.965 and 0.970, respectively) compared with no translation (T1 vs. T2). In addition, BiGAN and CycleGAN results are far from the ground truth values, that is, no translation (T1 vs. T1 and T2 vs. T2) is observed. The proposed GAN achieved lower BD (0.011) and higher HC values (0.997) than no translation (T1 vs. T2), BiGAN, and CycleGAN. In the case of T2 translation into T1, the proposed model achieves BD and HC values that are almost close to the ground truth values, that is, no translation (T1 vs. T1 and T2 vs. T2). From Table 2, it is clear that BiGAN and CycleGAN achieved low BD (0.195 and 0.152, respectively) but not high HC (0.771 and 0.795, respectively) compared with no translation (T1 vs. T2). In addition, BiGAN and CycleGAN results are far from the ground truth values, that is, no translation (T1 vs. T1 and T2 vs. T2). The proposed GAN achieved low BD (0.078) and high HC values (0.889) compared to BiGAN and CycleGAN, but the HC value of the proposed GAN was low compared to that of no translation (T1 vs. T2). This indicates that T1 modality translation into T2(Table  1) is more accurate and effective than T2 modality translation into T1(Table 2).

\subsection{Model training and Loss Consistency}

\begin{figure*}
\centering
{\includegraphics[scale=0.75]{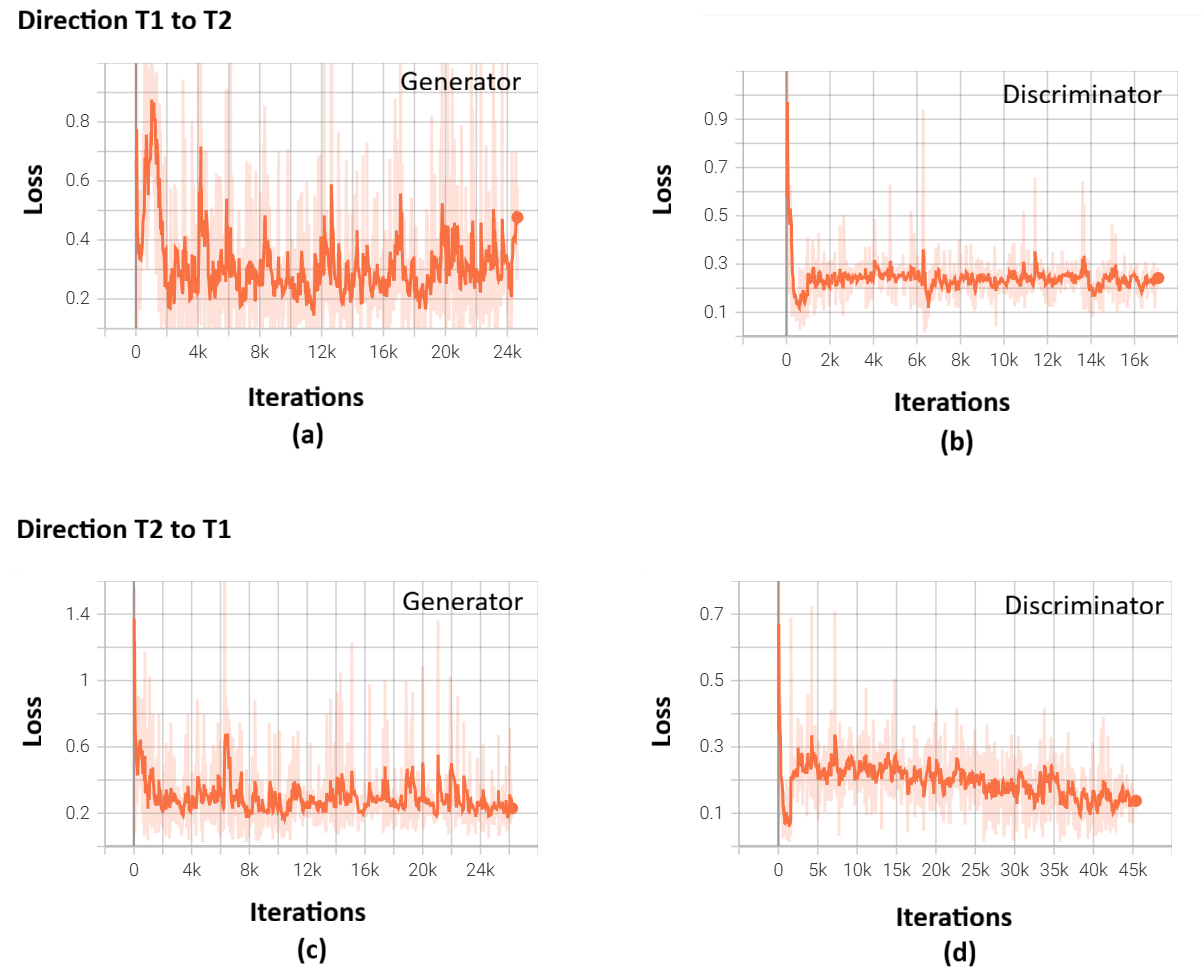}}
\caption{Generator and discriminator losses of BiGAN for both translation T1 into T2 and T2 into T1.}
\label{Figure3}
\end{figure*}

\begin{figure*}
\centering
{\includegraphics[scale=0.75]{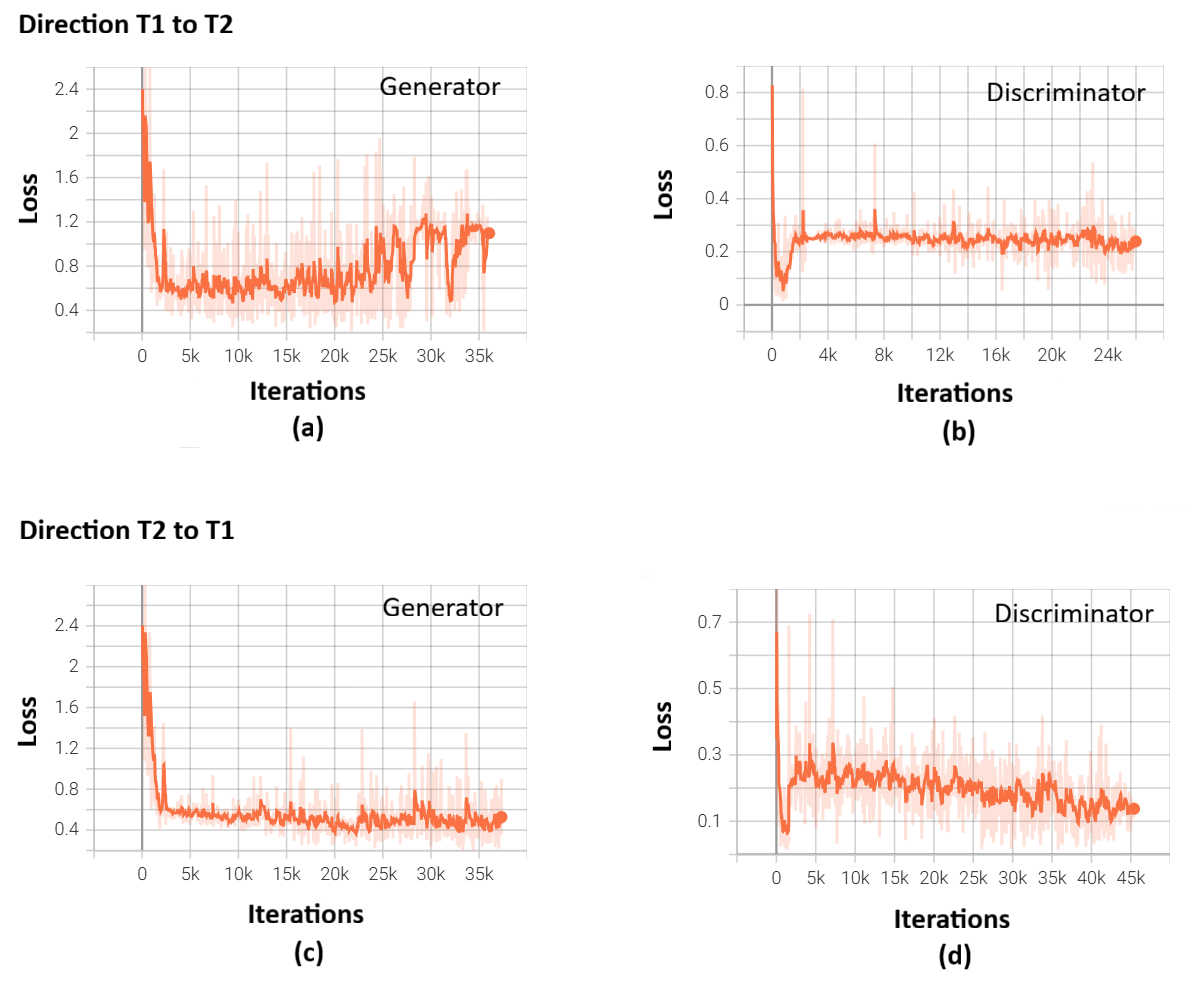}}
\caption{Generator and discriminator losses of CycleGAN for both translation T1 into T2 and T2 into T1. }
\label{Figure3}
\end{figure*}

\begin{figure*}
\centering
{\includegraphics[scale=0.75]{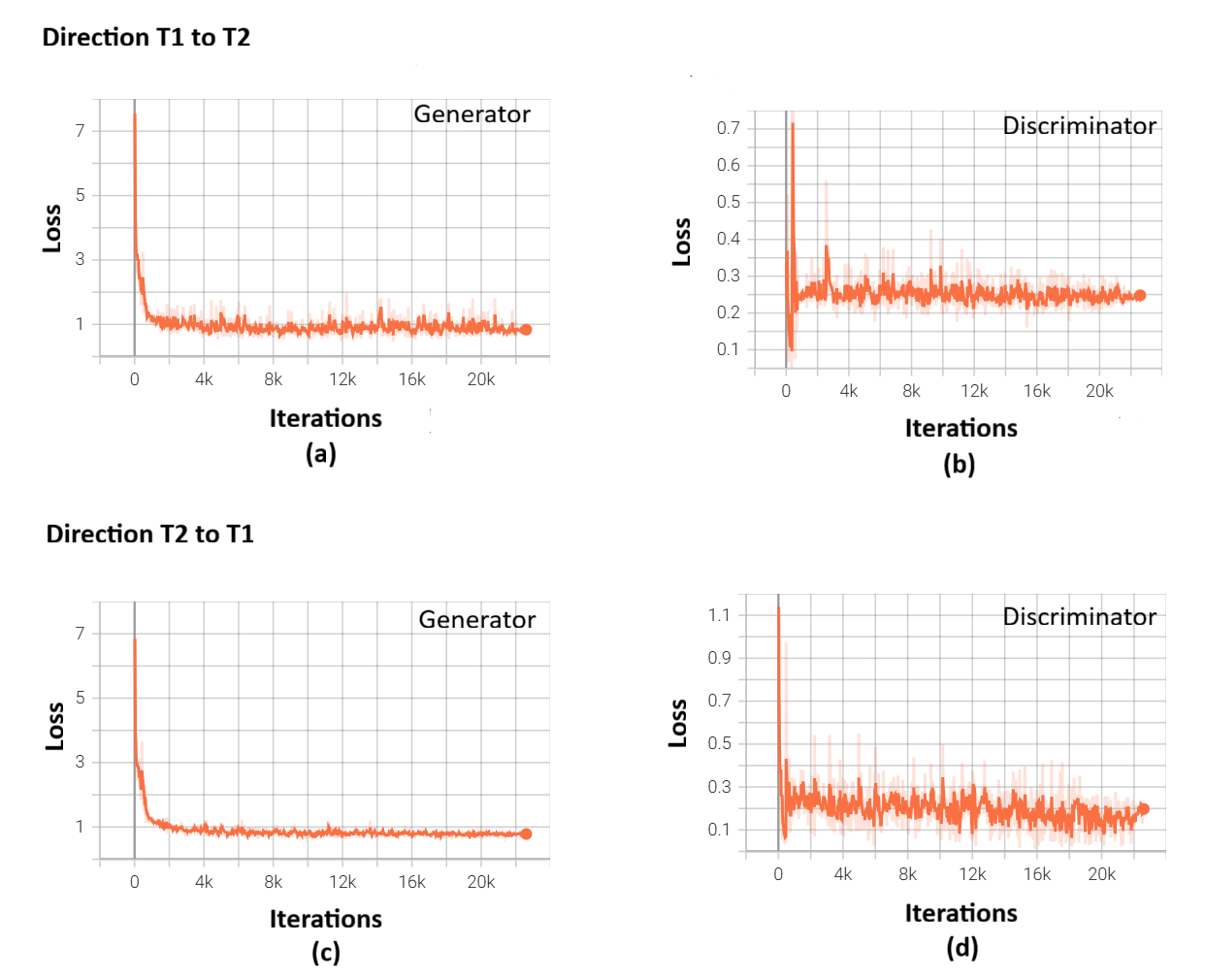}}
\caption{Generator and discriminator losses of Proposed GAN for both translation T1 into T2 and T2 into T1.}
\label{Figure3}
\end{figure*}

We visualized the generator and discriminator losses of all three models, including BiGAN, CycleGAN, and fine-tuned CycleGAN (fig 7, 8, and 9) to better understand the model training and its nature in the adaptation of the T1 modality into T2 and T2 modality into T1. In general, the generator loss curves decrease (see fig 7(a) and (c), 8(a) and (c), 9(a) and (c)), and the discriminator loss is consistent between 0.1 and 0.3 (see fig. 7(b) and (d), 8 (b) and (d), 9 (b) and (d)), which represents the fine training of all models. Moreover, the default CycleGAN generator losses (Figs.8 (a) and (c)) are slightly smoother and decrease more than those of BiGAN (Figs.7 (a) and (c)), and the fine-tuned CycleGAN (Figs.9 (a) and (c)) generator losses are much smoother and consistently decrease more than those of the default CycleGAN (Figs.8 (a) and (c)). This indicates that the fine-tuned CycleGAN model trained and performed better than BiGAN and CycleGAN. Furthermore, from the loss curve, it can be seen that all three models trained and worked better when the model translated the T2 modality into T1 (see fig. 7 (c), 8 (c), and 9 (c)) compared to the T1 modality translating into T2 (see fig. 7 (a), 8 (a), and 9 (a)).      

\section{Conclusion}
This study presents an effective strategy for the domain adaptation of MRI images using unpaired image-to-image translation. By leveraging a fine-tuned Cycle-GAN architecture over unpaired datasets and integrating disparity loss into the objective function, the proposed method achieved improved performance in synthesizing MRI modalities across one to another domains. Experimental results with BiGAN, default CycleGAN, and fine-tuned CycleGAN demonstrate that all approaches can successfully adapt or translate the T1 modality into T2 and the T2 modality into T1. Moreover, we evaluated the robustness of all three models using the SSIM map, scores, and visual image translation. The SSIM map and values between T1 and translated T1, and T2 and translated T2, are higher than those between T1 and T2 for all three models. Furthermore, the BD and HC values show that the fine-tuned CycleGAN model works much better than BiGAN and the default CycleGAN model. In future studies, we will further modify and test our fine-tuned CycleGAN model using other medical imaging datasets such as ultrasound and OCT. 

\paragraph{Declaration of generative AI and AI-assisted technologies in the manuscript preparation process:}
During the preparation of this work, the author(s) used Paper Pal and SciSpace to improve writing. After using this tool/service, the author(s) reviewed and edited the content as needed and took (s) full responsibility for the content of the published article.


\bibliographystyle{unsrt}  
\bibliography{arxiv}

\end{document}